\documentclass[journal]{IEEEtran}
\ifCLASSINFOpdf
\else
\fi
\usepackage{amssymb}
\usepackage{graphicx}

\begin{document}
%
\title{Wallpaper Texture Generation and Style Transfer Based on Multi-label Semantics}
%
%
%

\author{Ying~Gao,
        Xiaohan~Feng,
        Tiange~Zhang,
        Eric~Rigall,
        Huiyu~Zhou,
        Lin~Qi,
        and~Junyu~Dong
\thanks{Manuscript received November 6, 2020; revised April 1, 2021; accepted April 29, 2021. The work of Junyu Dong was supported in part by the National Key
Research and Development Program of China under Grant 2018AAA0100602, in part by the National Natural Science Foundation of China under Grant 41927805,
U1706218, and in part by the Natural Science Foundation of Shandong Province, China, under Grant ZR2018ZB0852. The work of Lin Qi was supported by the
National Natural Science Foundation of China under Grant 61501417. The work of Huiyu Zhou was supported in part by the Royal Society-Newton Advanced
Fellowship under Grant NA160342, in part by the U.K. Engineering and Physical Sciences Research Council (EPSRC) under Grant EP/N011074/1, in part by the
Royal Society-Newton Advanced Fellow-ship under Grant NA160342, and in part by the European Union's Horizon 2020 Research and Innovation Program through
the Marie-Sklodowska-Curie under Grant 720325. This article was recommended by Associate Editor L. Marcenaro. (Corresponding author: Junyu Dong.)}
\thanks{Ying Gao, Xiaohan Feng, Tiange Zhang, Eric Rigall, Lin Qi and Junyu Dong are with the College of Computer Science and Technology, Ocean University of China,
Qingdao 266071, China (e-mail: dongjunyu@ouc.edu.cn).}
\thanks{H. Zhou is with University of Leicester, United Kingdom.}
\thanks{Color versions of one or more figures in this article are available at}
\thanks{https://doi.org/10.1109/TCSVT.2021.3078560.}
\thanks{Digital Object Identifier 10.1109/TCSVT.2021.3078560.}}

%
%

\markboth{IEEE TRANSACTIONS ON CIRCUITS AND SYSTEMS FOR VIDEO TECHNOLOGY}%
{Shell \MakeLowercase{\textit{et al.}}: Bare Demo of IEEEtran.cls for IEEE Journals}
%



\maketitle

\begin{abstract}
Textures contain a wealth of image information and are widely used in various fields such as computer graphics and computer vision. With the development of machine learning, the texture synthesis and generation have been greatly improved. As a very common element in everyday life, wallpapers contain a wealth of texture information, making it difficult to annotate with a simple single label. Moreover, wallpaper designers spend significant time to create different styles of wallpaper. For this purpose, this paper proposes to describe wallpaper texture images by using multi-label semantics. Based on these labels and generative adversarial networks, we present a framework for perception driven wallpaper texture generation and style transfer. In this framework, a perceptual model is trained to recognize whether the wallpapers produced by the generator network are sufficiently realistic and have the attribute designated by given perceptual description; these multi-label semantic attributes are treated as condition variables to generate wallpaper images. The generated wallpaper images can be converted to those with well-known artist styles using CycleGAN. Finally, using the aesthetic evaluation method, the generated wallpaper images are quantitatively measured. The experimental results demonstrate that the proposed method can generate wallpaper textures conforming to human aesthetics and have artistic characteristics.
\end{abstract}


%
\IEEEpeerreviewmaketitle

\section{Introduction}
\label{sec:intro}
Textures can be used to express the features and structural levels of an object's surface, including rich image feature information, and are therefore widely used in many image processing tasks. The texture within an image refers to the difference in color, pattern, or illumination, causing the image to appear in structure and space with regular or cluttered details. As one type of image, wallpaper images also contain rich textures. People choose beautiful wallpaper to decorate their rooms, which not only enhances the impression of rooms, but also brings visual pleasure.

Wallpaper designers need to collect a large amount of materials to create a new wallpaper. However, the materials of related styles that can be collected are often limited, so designers have to draw relevant patterns themselves. This is a very strenuous and time-consuming job, to such an extent that designers require a quick method for producing different styles of wallpaper to assist their work. The current research lacks relevant methods, and the wallpaper data with annotations is also limited.

In the field of computer vision, researchers annotate images with labels. Traditional supervised learning mainly uses a training set to learn a mapping function for a single label. Wallpaper texture images contain a wealth of texture information that is difficult to annotate with a simple single label; thereby we propose a multi-label semantic mechanism for wallpaper texture using multi-label semantics. Through wallpaper description collection experiments, we first obtain the semantic labels for a small part of the data; then we extract the features from wallpaper texture images, and learn the label distribution and feature vectors of the labeled image. This process can predict the label distribution for unlabeled images and label new images, providing a way to augment the labeled dataset.

Traditional texture generation methods are mainly based on mathematical models, and the generated texture images are formulaic and rigid. As wallpaper texture features vary in colors and textures, it is difficult to generate appealing wallpaper textures using mathematical models. In 2006, Hinton et al. \cite{Hinton2006Reducing} proposed a deep learning method, which was inspired by the biological structure and operational mechanism of the human brain, and was widely used in popular fields such as speech recognition, image processing, and natural language processing. Inspired by their work, the research on image generation has also made rapid progress. One of the most representative work is the Generative Adversarial Network (GAN) proposed by Goodfellow et al. \cite{Goodfellow2014Generative} in 2014. This network consists of two competing neural network models: one is called the generator, taking noise as input and learns how to generate samples similar to ground truth data, and the other network is called the discriminator, receiving generator and ground truth data, and being trained as a classifier to correctly distinguish them. GAN has achieved good results in many tasks such as image synthesis and generation, but when used for wallpaper texture generation, the components and spatial layout of generated texture images are not well controlled, and the textons are constantly too similar or their distributions too cluttered.

In order to solve the abovementioned problems, in this paper, we propose a wallpaper texture image generation model based on multi-label semantics, and we improve the wallpaper dataset with multiple styles and semantic description labels, using perception-driven texture generation technique in the view of generating wallpaper images, that meet user requirements according to the user's perceptual descriptions. This model then uses a style transfer method to convert the style of wallpaper images, and assist the artistic creation work from an artist perspective. Our methods can generate artistically-inspired wallpaper images and aesthetically evaluate the resulting images. The proposed framework has application value in the research fields of computer vision and image processing, and also has great application prospects in areas of designing, such as artistic designing.

The contributions are as follows: We improve the wallpaper dataset with multiple styles and semantic description labels. We proposed a perception-driven texture generation method for wallpaper texture generation with multi-labels semantics, and achieved good experimental results according to the user's perceptual descriptions. On the basis of previous work, we further optimized and tested the generation model using different perceptual models, and compared the generation results obtained using different image features. We further processed the generated wallpaper texture using different style transfer methods, and aesthetically evaluate the resulting images.

The rest of this paper is organized as follows. In Section \ref{sec:related}, we review the related work in the area of multi-label semantics, texture generation and style transfer. We introduce the multi-label semantic descriptions of the designed wallpaper dataset in Section \ref{sec:dataset}. In Section \ref{sec:method}, we detail the proposed framework of label prediction, wallpaper texture generation, style transfer and aesthetic quality assessment. The experimental results are discussed in Section \ref{sec:experiments} and the ablation experiments are shown in Section\ref{sec:Ablation}. Finally, in Section \ref{sec:conclusion}, we give conclusions and discuss the future work.

\section{Related Work}
\label{sec:related}
\subsection{Semantic descriptions}
Semantic descriptions are determinant to find structure in texture data. As early as 1978, Tamura et al. \cite{Hideyuki1978Textural} proposed to use a more understandable language to represent texture features and summarized them into six texture semantic feature components. Schwartz et al. \cite{Schwartz2013Visual} recently proposed to describe the texture properties of different materials using visual material properties, such as metallic and cortical, encoded into the appearance of material properties with their corresponding characteristics. In single-label learning algorithms, an instance has only one category label, that is, only one semantic descriptor. However, the sample instances in real life are complex and varied, including multiple elements. Using only one descriptor is not enough to completely represent a texture image. Multiple descriptors are needed to characterize the sample as completely as possible. Giunchiglia et al. \cite{ Giunchiglia2020Coherent} recently proposed C-HMCNN(h) which exploits the hierarchy information in order to produce predictions coherent with the constraint. Lin et al. \cite{Lin2018End} proposed $ E^2FE $ which directly learns a feature-aware code matrix via jointly maximizing the recoverability of the label space and the predictability of the latent space, and gains performance improvements over other state-of-the-art LSDR methods.

In 2016, Geng et al. \cite{Xin2016Label} proposed a multi-label learning method named Label Distribution Learning algorithm (LDL). In traditional multi-label problems, an instance is described by multiple labels, the label distribution should not only consider the use of multiple descriptive labels, but also the degree related to each label. The LDL algorithm, formulating the label distribution learning problem, is a broader application framework, which has achieved promising results on many computer vision tasks, including the emotion distribution of facial expressions \cite{Zhou2015Emotion}, facial age estimation \cite{Geng2013Facial}, head pose estimation \cite{Geng2014Head}, prediction of crowd opinion on movies \cite{Geng2015Prerelease} and so on.

\subsection{Texture generation}
In earlier studies, texture synthesis were procedural and sample-based \cite{Efros1999Texture}. With the development of deep learning, novel methods based on convolutional neural network have been proposed. Vacher et al. \cite{Vacher2020Texture} proposed natural geodesics arising with the optimal transport metric to interpolate between arbitrary textures. Zhao et al. \cite{Zhao2019Pixelated} proposed to exploit pixelated object semantics to guide image colorization,and produced more realistic and finer results. Luan et al. \cite{Luan2018Gabor} proposed to incorporate Gabor filters into DCNNs to enhance the resistance of deep learned features to the orientation and scale changes, which have much fewer learnable network parameters. By using the convolutional layers and fully-connected layers to deepen the networks, the fine-grained features from images can be extracted, yielding good results in tasks such as image classification and semantic retrieval \cite{Lecun1998Gradient,Alex2012ImageNet,Karen2014Very,Kaiming2015Deep}.

However, the fully connected network is computationally difficult to train even by the gradient descent method, therefore the texture generation work is not suitable when using a fully connected network. On this basis, Goodfellow et al. \cite{Goodfellow2014Generative} proposed Generative Adversarial Networks (GAN). The generator and discriminator models included in this network are continuously optimized during the training process, and finally converge to a stable state: the generator model is capable of generating sufficiently realistic images, and the discriminant model cannot distinguish between generated and real images. The first version of GAN faced many learning issues, that researchers tackled from many aspects. Among them, unstable training often led the generator produce meaningless output. Radford et al. \cite{Radford2015Unsupervised} proposed a deep convolutional Generative Adversarial Network, which has certain architectural constraints, making the network robust to unsupervised learning. The Conditional Generative Adversarial Nets (CGAN) \cite{Mirza2014Conditional} added constraints during the training of GAN, turned unsupervised training into semi-supervised or supervised learning. LAPGAN \cite{Denton2015Deep} used the Laplacian pyramid to implement the serialization learning process, combined with the idea of residual learning, reducing the difficulty of learning.

GAN-based image synthesis methods have produced good results in many fields, such as facial age synthesis, Liu et al. \cite{Liu2017Face} proposed C-GANs with a conditional transformation network and two discriminative networks, age discriminative network and transition pattern discriminative network, which produces good performance in face aging. In 2020, Sun et al. \cite{Sun2020Facial} proposed a label distribution-guided generative adversarial network based on GANs, which can well capture the correlation among different age groups, so that smooth aging sequences can be achieved. Despite the research development around GANs, there still remain limitations for applying them to wallpaper synthesis and texture generation. Indeed, the generating process of a new wallpaper, only based on random noise, is uncontrollable, the components and spatial layout of the generated texture images are not well controlled, and the textons are always too similar or their distributions too cluttered. Therefore, in this paper, we propose a perception driven wallpaper texture image generation model based on multi-label semantics, generating wallpapers based on semantic perception.

\subsection{Style Transfer}
Style transfer, as one of the most artistic and creative research topics in the field of deep learning, has recently attracted a lot of attention. Gatys et al. \cite{Gatys2015Neural,Leon2016Image} proposed to introduce deep learning into style learning, and to obtain a new image by incorporating the style of an image into another image. The new image combines the style of the first image with the content of the second. In 2016, Johnson et al. \cite{Johnson2016Perceptual} proposed the use of perceptual loss functions for training feed-forward networks for image transformation tasks. In \cite{Li2016Combining}, the authors studied the combination of Markov random field models and convolutional neural networks to improve the quality of the generated images. Image aesthetic quality assessment is one of the novel research work in the field of computer vision. Traditional aesthetic quality assessment includes judging texture, shape, color and depth of field. \cite{Datta2006Studying,Ke2016The}. Luo et al. \cite{Luo2008Photo} separated the foreground and background of images during the aesthetic evaluation, and took the contrast between the two as an important feature. Wong et al. \cite{Wong2010Saliency} extracted the foreground using the saliency detection method. In 2017, the Google team proposed NIMA \cite{Talebi2017NIMA}, an assessment method to predict the distribution of human ratings on images from an aesthetic perspective, providing satisfactory results. Therefore, in this paper, the NIMA model is used to evaluate the generated wallpaper texture image.

In our work, we propose a wallpaper texture image generation model, and we improve the wallpaper dataset with multiple styles and semantic description labels. The proposed method can generate wallpapers according to the user's perceptual descriptions, and the generated wallpapers are further applied in style transfer and aesthetic evaluation, which has application value and application prospects in computer vision and designing.

\section{Wallpaper texture dataset}
\label{sec:dataset}
\begin{figure*}
\centering
  \includegraphics[width=0.98\textwidth]{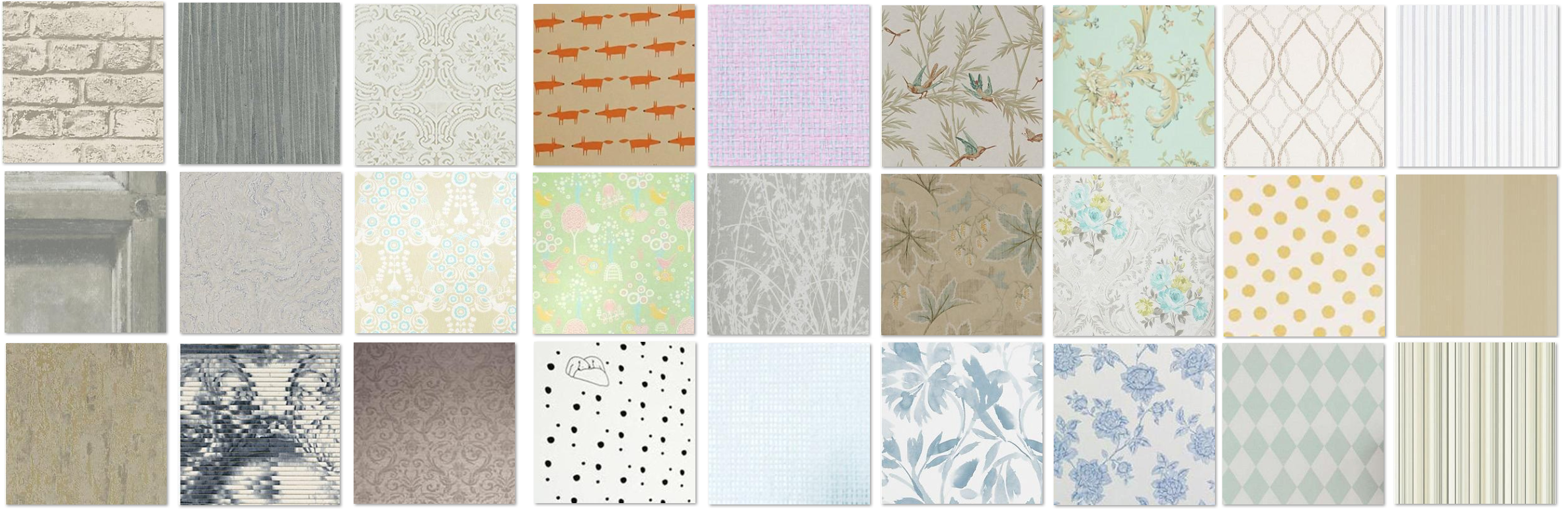}
\caption{Nine categories of wallpaper images in the database, from left to right, the keywords of the style are vintage, post-modern, classical, fresh, modern, country-style, flora, geometric and striped. }
\label{fig:wallpaper}       
\end{figure*}

In order to generate wallpapers, firstly, we need a database of wallpapers. In our previous work \cite{Feng2018Predicting} we built a wallpaper dataset, it includes nine types of wallpaper images collected from two online wallpaper sale websites \cite{wallpaperfromthe70s,houzz}, for a total of 1800 wallpaper images. The first six categories are defined by their styles, whereas the last three categories are mainly described by their content or contained elements. According to the wallpaper dataset introduction, the keywords for style are as follow: vintage, post-modern, European classical(or classical), fresh, modern, country-style, flora, geometric and striped. Fig.~\ref{fig:wallpaper} shows the different styles of wallpaper images in the database.

A single style label cannot accurately describe the wallpaper texture image. Thus, we conducted user survey to capture the visual perception of the participants when observing the wallpaper texture dataset images, then collected their perceptual descriptions, and finally obtained multi-label semantic descriptions of the wallpaper texture dataset.

Thirty-nine participants joined this wallpaper description collection survey, they were asked to view wallpaper pictures one by one, then use appropriate words to describe the images they see from a predefined list of words, and score each word according to their own feelings, ranging in [1, 100]. This means that if ones feels a word can strongly describe the wallpaper they see, score highly; conversely, if one thinks a word is not appropriate, with only a small correspondance to the wallpaper, score lowly. The description of each image should be consistent with the participants' own feelings. The requirement is that, giving the descriptive words, the image formed in the brain can be consistent with the image shown in the screen. Some descriptive words used frequently are provided just for reference only, and the participants can use their descriptive words to describe as long as they feel appropriate.

We count the descriptive words and find the ones that are frequently used. Meanwhile, we omit the words that are used less or unsuited for wallpaper labels. Finally, we choose 93 words as wallpaper labels, including 38 adjectives and 55 nouns. Because color is important to the style of wallpaper, we add a color value as wallpaper label. There are 94 descriptive labels in all.

Adjectives include: repetitive, floral, regular, fresh, vintage, wooden, blurry, striped, zig-zag, wavy, classical, undertint, simple, modern, lovely, worn-out, elegant, serried, exquisite, patterned, symmetrical, country-style, bright-colored, complex, bright, post-modern, colorful, messy, somber, granular, cracked, coarse, spotted, marble, stone, bent, netted, geometric.

Nouns include: triangle, leafage, rhombus, square, animal, circle, plant, bird, butterfly, dragonfly, tree, branch, letter, star, brick, plume, figure, polygon, book, bookshelf, grape, pinecone, pineapple, cherry, fish and grass, photo frame, automobile, bicycle balloon, building, cloud, airplane, mountain, arrow, mushroom, musical note, boat, horologe, dandelion, crown, button, robot, sky, vine, bowknot, seawater, water-drop, soil, skirt, snow, cotton, tyre, shoes, glasses, windmill.

After the experiment, referring to the glossary describing the texture summarized by Rao et al. \cite{Bhushan1997The}, we set descriptive words as wallpaper labels and calculate the value of each wallpaper label according to the scores subjects gave during experiment. For the adjective label, we add the scores of the same label, then normalized the value to $[0, 1]$. The value of adjective labels which is not used by subjects is set to $0$. For noun labels, we take another strategy, the value of noun labels is $0$ or $1$. Once a subject uses this noun label and the noun descriptor does exist in the wallpaper for sure, the value of the noun label is $1$, and $0$ otherwise. Finally, we compiled a 94-dimensional label as the semantic description of the wallpaper texture image.

\section{Method}
\label{sec:method}
\subsection{Wallpaper labels prediction}
A novel label learning algorithm, called label distribution learning (LDL)\cite{Xin2016Label}, which means learning on the instances labeled by label distributions. We use $x$ to represent instance variable, the $i-th$ instance is denoted by $x_i$, and we use $y$ to represent the label, the $j-th$ label value is denoted by $y_j$. The description degree of $y$ to $x$ is represented by $d^y_x$ and the label distribution of $x_i$ is represented by $D=\{d^{y1}_{xi},d^{y2}_{xi},...,d^{yc}_{xi}\}$, where $c$ is the number of possible label values. Labelling an instance $x$ is to assign a real number $d^y_x$ to each possible label $y$, representing the degree to which $y$ describes $x$. Moreover, $d^y_x$ should meet two conditions: $d^y_x\in[0,1]$, and $\Sigma_y{d^y_x}=1$. \cite{Xin2016Label} proposed six LDL algorithms in three ways: problem transformation, algorithm adaptation, and specialized algorithm design. In order to compare these algorithms, we use six evaluation measures to compare all algorithms, Table.~\ref{tab:measure} lists the formulae of the six measures. The $\uparrow$ means the larger the better, and the $\downarrow$ means the smaller the better.

\begin{table*}
  \centering
  \caption{Evaluation measure for the distribution distance/similarity measures.}\label{tab:measure}
  \begin{tabular}{ll}
  \hline\hline
  \ Measure   &Formula  \\ \hline
   Chebyshev$\downarrow$ &$Dis_1(D,\widehat{D})=\max|d_j-\widehat{d}_j|$  \\
   Clark$\downarrow$& $Dis_2(D,\widehat{D})=\sqrt{\sum^c_{j=1}\frac{(d_j-\widehat{d}_j)^2}{(d_j+\widehat{d}_j)^2}}$    \\
   Canberra$\downarrow$& $Dis_3(D,\widehat{D})=\sum^c_{j=1}\frac{|d_j-\widehat{d}_j|}{d_j+\widehat{d}_j}$    \\
   Kullback-Leibler$\downarrow$& $Dis_4(D,\widehat{D})=\sum^c_{j=1}d_j\ln\frac{d_j}{\widehat{d}_j}$    \\
   Cosine$\uparrow$&  $Sim_1(D,\widehat{D})=\frac{\sum^c_{j=1}d_j\widehat{d}_j}{\sqrt{\sum^c_{j=1}{d_j}^2}\sqrt{\sum^c_{j=1}{\widehat{d}_j}^2}}$   \\
   Intersection$\uparrow$& $Sim_2(D,\widehat{D})=\sum^c_{j=1}\min(d_j,\widehat{d}_j)$    \\ \hline\hline
  \end{tabular}
\end{table*}

Label distribution has a data form similar to probability distribution and shares the same conditions. We can use the form of conditional probability to represent $d^y_x$,i.e. $d^y_x=P(y|x)$. Suppose $P(y|x)$ is a parametric model $P(y|x;\theta)$, where $\theta$ is the parameter vector. Given the training set $S$, LDL's target is to find the $\theta$ that can generate a distribution similar to the distribution given the instance $X_i$. As an example, the KL divergence is used as the distance measure method, then then the best parameter vector $\theta^*$ is determined as:
\begin{eqnarray}
\theta^* &=&\arg \min_\theta\sum_{i}\sum_{j}(d^{y_j}_{x_i}\ln\frac{d^{y_j}_{x_i}}{p(y_j|x_i;\theta)}) \nonumber \\
&=& \arg \max_\theta \sum_{i}\sum_{j}(d^{y_j}_{x_i}\ln p(y_j|x_i;\theta))
\end{eqnarray}
For SLL, $d^{y_j}_{x_i}=Kr(y_j,y(x_i))$, where $Kr(.,.)$ is the Kronecker delta function and $y(x_i)$ is the single label to $x_i$. Eq (1) can be changed to:
\begin{eqnarray}
\theta^* &=& \arg \max_\theta \sum\ln p(y|x_i;\theta)
\end{eqnarray}
For MLL, each instance is labeled with a label set, and then Eq (1) is changed to:
\begin{eqnarray}
\theta^* &=& \arg \max_\theta \sum_{i}\frac{1}{|Y_i|}\sum_{y\in Y_i}\ln p(y|x_i;\theta)
\end{eqnarray}

\subsection{Perception driven wallpaper texture generation}
Texture synthesis and generation have been studied for years, and are still hot topics recently. Many effective methods were proposed for texture synthesis, such as pixel-based \cite{Paget1998Texture} and patch-based \cite{Efros1999Texture}. With the development of deep learning, novel approaches of texture generation have become popular. GANs \cite{Goodfellow2014Generative} are a recent approach to train generative models of data, which have been shown to work particularly well on image data. Based on GAN, Mirza et al. \cite{Mirza2014Conditional} proposed a conditional Generative Adversarial Net (CGAN) in 2014. CGAN adds some constraints to the GAN network, and introduces condition variables in both the generator network and the discriminator network. Inspired by CGAN and on the basis of our preliminary work \cite{Gan2017Perception}, this paper takes multi-label semantic attributes as condition variables to generate wallpaper textures, and constructs a perception driven generation model for wallpaper texture image generation.

In order to perform perceptually driven wallpaper texture generation based on multi-label semantics, it is first necessary to train a deep neural network for perceptual feature regression, called a perceptual model, to semantically constrain the generated wallpaper. The perceptual model here uses the modified Inception-v3 model \cite{Szegedy2016Rethinking}. Inception-v3 was originally used for object recognition, which is a classification task. Our multi-label semantics follow a [0,1] distribution; accordingly, in order to perform perceptual semantic regression, we modify the network structure of Inception-V3. The overall loss function of the original network is composed of the weighted sum of the auxiliary output and the softmax cross-entropy of the final output. In order to enable the network to perform perceptual semantic regression, this paper changes the activation function of the auxiliary output and final output in the original network from softmax to hyperbolic tangent function, and changes the definition of the network loss function from cross entropy loss to quadratic loss. During the construction of the joint model, the perceptual model needs to be pre-trained first, and then fixed as a part of the jiont model. Here $H$ is used to represent the perceptual model, $x$ and $y$ represent wallpaper images and semantic vectors, respectively, and the pre-training process of the perceptual model can be expressed as:
\begin{eqnarray}
\min_H \frac{1}{2}E_{x:P_{data}(x)}[(h(x)-y)^2]
\end{eqnarray}

\begin{figure*}
\centering
  \includegraphics[width=0.98\textwidth]{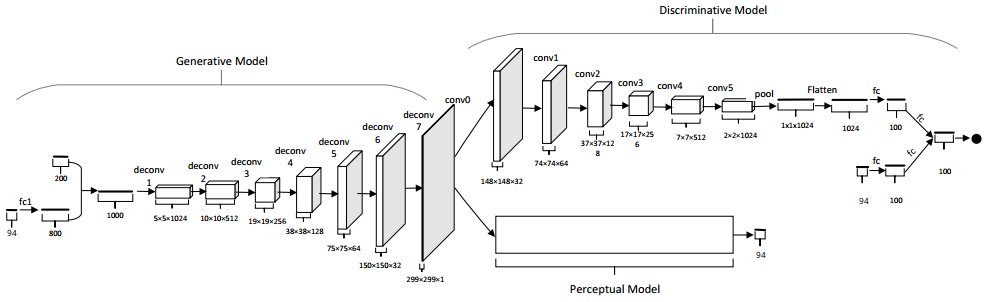}
\caption{The architecture of joint model for perception driven wallpaper texture generation based on multi-label semantics. }
\label{fig:architecture}       
\end{figure*}

This perceptual model is used to recognize if the wallpapers produced by the generator network are sufficiently realistic and have a given perceptual description. We assume that the generated wallpaper image is real enough and has a given semantic description; then it should be able to be correctly recognized by the perceptual model. Therefore, we use the perceptual model to impose perceptual constraints on the generative model, imitating the way humans work when drawing images. Moreover, the perceptual model also aids the discriminator network in differentiating inputs, increasing the discriminator network attention on determining whether a wallpaper image is from real instance space. Thus, the discriminator network is mainly responsible for judging the authenticity of the generated wallpaper texture images, and the perceptual model ensures that the generator network produces wallpaper texture images conform to the perceptual feature description. In this way, the wallpaper images generated by the generative model can have the desired semantic attributes while maintaining authenticity. Fig.~\ref{fig:architecture} illustrates the architecture of joint model for perception driven texture generation.

In this paper, the perceptual model loss is modified, and the cosine loss is added. The cosine similarity is obtained by calculating the cosine of the angle between two vectors. The closer its value to 1, the higher the similarity between the two vector distributions, and conversely, the closer its value to 0, the lower the similarity. After pre-training the perceptual model, the latter is fixed as part of the joint model. Here, $F$ is used to represent the perceptual model. The loss function of the perceptual model is:
\begin{eqnarray}
F_{loss} &=& \frac{1}{2n}\sum_{i=1}^{n}{(F(x_i)-y_i)}^2  + \nonumber\\
         & & \frac{\sum_{i=1}^n(F(x_i)\times y_i)}{\sqrt{\sum_{i=1}^n(F(x_i))^2}\times \sqrt{\sum_{i=1}^n(y_i)^2}}
\end{eqnarray}
Where $x_i$ is training example, $y_i$ is the corresponding semantic feature vector, $n$ is the number of training instances. After completing the training of the perceptual model, the joint model is trained. The loss of discriminative model is defined as:
\begin{eqnarray}
D_{loss} &=& -\frac{1}{n}\sum_{i=1}^{n}(q_i\log D(x_i.y_i)) + \nonumber\\
         & & (1-q_i)\log(1-D(x_i,y_i))
\end{eqnarray}
Where $q_i$ is 1 or 0, which corresponds respectively to real pair $(x_i, y_i)$ or not. The loss of generative model includes both the loss from discriminative model and the loss from perceptual model. The generative network not only makes the generated wallpaper image as realistic as possible, but also makes its perceptual attributes as consistent as possible with the existing perceptual features in the database. More precisely, the loss of generative model is defined as:
\begin{eqnarray}
G_{loss} &=& G_{loss\_d}+\alpha \ast G_{loss\_f}
\end{eqnarray}
\begin{eqnarray}
G_{loss\_d} &=& - \frac{1}{n} \sum_{i=1}^n \log(D(G(y_i,z_i),y_i))  \\
G_{loss\_f} &=& \frac{1}{2n} \sum_{i=1}^n (F(G(y_i,z_i))-y_i)^2
\end{eqnarray}
where $\alpha$ is a tradeoff parameter, $z_i$ is a random noise vector. Perceptual model is preliminarily trained, and both the generative and discriminative models are trained in an adversarial manner. So that, the discriminative model makes the generator produce realistic textures, while the perceptual model makes the generated textures possess certain semantic attributes. During the training process of the joint model, the discriminant model is optimized first, and then the generative model is optimized twice in succession. The perceptual model remains the same during the training process of the joint model, and it is used to provide perceptual supervision to the generative model and perceptually evaluate the generated wallpaper images. During the model training process, a total of 316,000 iterations were performed, the Adam algorithm is used for training, and $\alpha$ was set to 10.

\subsection{Wallpaper texture style transfer}
People choose wallpapers without special requirements on the style. They tend to like wallpapers with a variety of styles and elements. As one of the most recent artistic and creative research topics in the field of deep learning, style transfer has attracted a lot of attention.
In many cases, unlike for two styles of images, collecting two images with the same content and different styles is hard. Therefore, it is tricky to train the network with unpaired data to perform style transfer. In this paper, the wallpaper texture style transfer is realized based on CycleGAN \cite{Zhu2017Unpaired}. CycleGAN allows images from two different domains to convert to each other in the absence of paired examples. CycleGAN trains two GAN networks simultaneously, one GAN network transforms from domain A to domain B and the other GAN network transforms from domain B to domain A, which realizes a bidirectional mapping. The two GAN networks have each a discriminator, but share parameters in the two generators, therefore the model totalizes two generators, two discriminators, and four losses.

In this paper, we use CycleGAN to convert generated wallpaper images with our database semantics into wallpapers with some well-known artist styles. In order to train models with these styles, we collected a lot of artists' works from an art works website \cite{wikiart}, including Vincent Van Gogh (Dutch Impressionist painter), Paul Cezanne (French post-impressionist painter), Paul Gauguin (French post-impressionist painter), Henri Matisse (French painter, Fauvist founder), Pyotr Konchalovsky (Russian painter), Maurice Prendergast (American Impressionist landscape painter), Claude Monet (French Impressionist painter) and ukiyoe artists (Japan's Ukiyo-e). Different models can be trained using different style's works, and the pre-trained models are used to generate new wallpaper images consistent with their styles.

\section{Experiments}
\label{sec:experiments}
\subsection{Wallpaper labels prediction}
\begin{figure*}
\centering
  \includegraphics[width=0.98\textwidth]{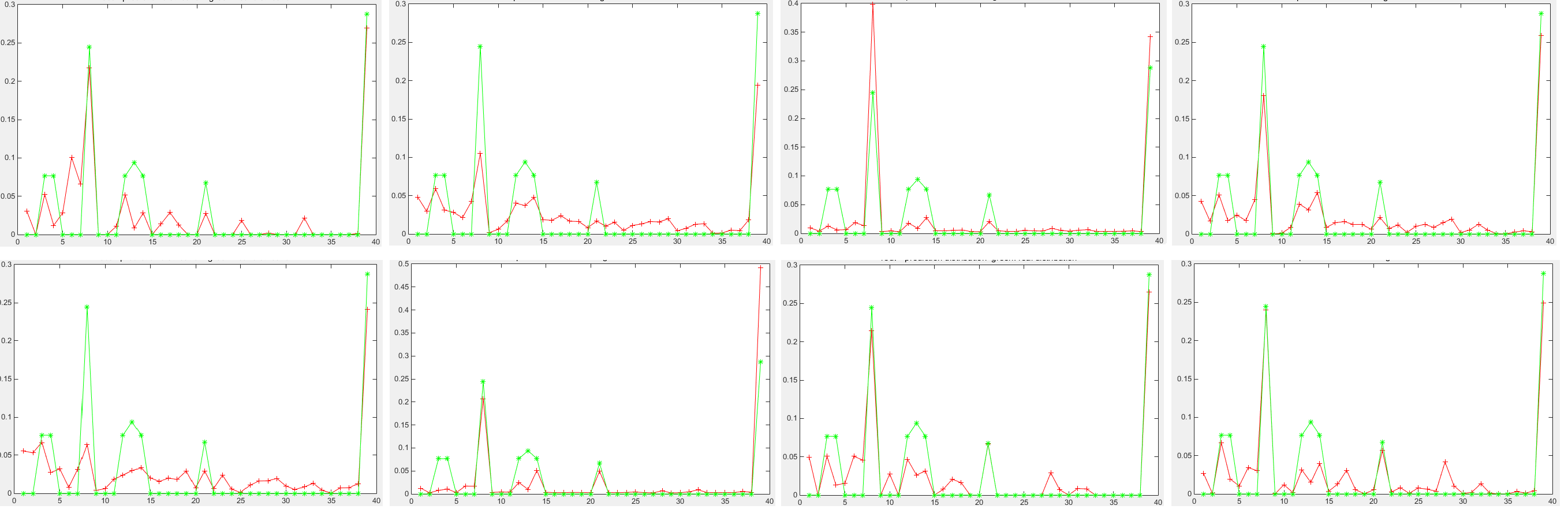}
\caption{The label distribution of one wallpaper texture. The x-axis represents the semantic labels, the y-axis indicates the value of each label. Combine each point to form the label distribution of a wallpaper. The red line is prediction distribution, and the green line is real distribution. From top to bottom, the feature of wallpaper images is extracted using gabor filter and alexnet, called gabor feature and deep feature, respectively. From left to right in the top line, the LDL algorithms are AA-knn, SA-IIS, SVR, and SA-BFGS, respectively. In the second line, from left to right, the LDL algorithms are AA-BP, SVR, AA-knn and SA-IIS respectively.}
\label{fig:ldl}       
\end{figure*}

We divided the 1800 wallpaper images into a training set (1500 images) and a test set (300 images). Experimental results of LDL are shown in Fig.~\ref{fig:ldl}. Table.~\ref{tab:gabor} and Table.~\ref{tab:deepfeature} list the average distance and average similarity between real and predicted label distribution of all testing samples using Gabor feature and deep feature respectively. The algorithms proposed in \cite{Xin2016Label} perform better than other modified algorithms.
\begin{table*}
  \centering
  \caption{Average distance and similarity measures between real and predicted distribution using gabor feature.}\label{tab:gabor}
  \begin{tabular}{ccccccc}
  \hline\hline
  \   &Chebyshev  &  Clark   &Canberra    &Kldist    &Consine    &Intersection\\ \hline
  AA-\textsl{k}NN&0.1638 & \verb|\| & \verb|\| &0.6688 &0.8260 &0.5696   \\
  SA-IIS&0.2146 &5.7681 &34.7082 &1.1278 &0.8233 &0.4544  \\
  SVR&0.2207 &5.7162 &34.7239 &0.8062 &0.7570 &0.5563  \\
  SA-BFGS&0.1874 &5.7570 &34.5226 &0.9770 &0.6994 &0.4741  \\ \hline\hline
  \end{tabular}
\end{table*}

\begin{table*}
  \centering
  \caption{Average distance and similarity measures between real and predicted distribution using deep feature.}\label{tab:deepfeature}
  \begin{tabular}{ccccccc}
  \hline\hline
  \   &Chebyshev  &  Clark   &Canberra    &Kldist    &Consine    &Intersection\\ \hline
  AA-BP&0.2236 &5.7771 &34.8115 &1.2983 &0.5970 &0.4230   \\
  SVR&0.1540 &5.7047 &33.7765 &0.4002 &0.8905 &0.7131  \\
  AA-\textsl{k}NN&0.1402 &\verb|\| &\verb|\| &0.5106 &0.8695 &0.6356  \\
  AA-IIS&0.1232 &5.6915 &33.7745 &0.4520 &0.8964 &0.6610  \\ \hline\hline
  \end{tabular}
\end{table*}

\begin{figure}
\centering
  \includegraphics[width=0.35\textwidth]{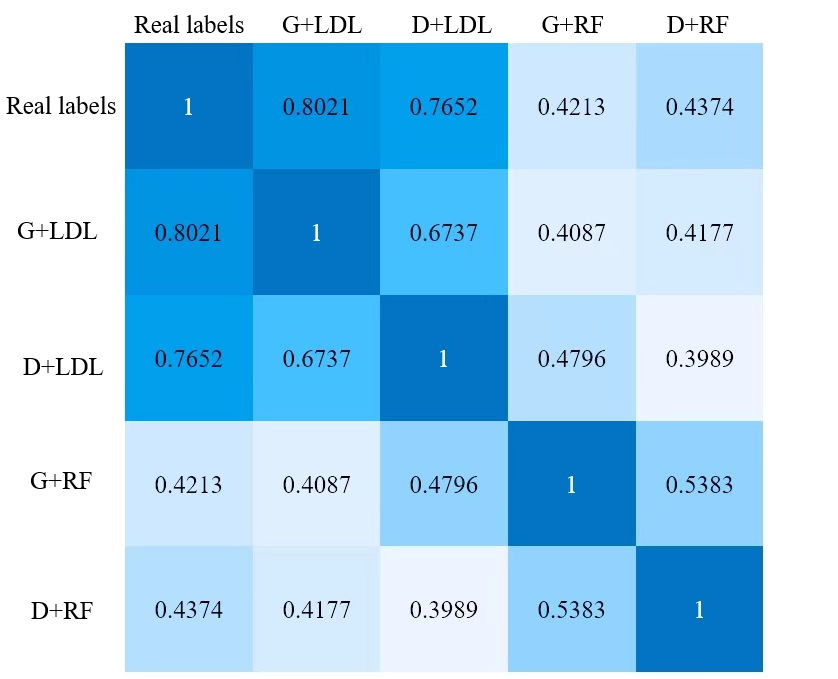}
\caption{Correlation coefficient between the predicted labels and the real labels. From left to right, from top to bottom, they are the real labels, the labels predicted by Gabor features and label distribution learning, the labels predicted by deep feature and label distribution learning, the labels predicted by Gabor features and regression Random Forest, and The deep feature and the label predicted by the regression Random Forest.}
\label{fig:corr}       
\end{figure}

In order to further compare about multi-label semantics for wallpaper dataset evaluation, we use regression Random Forest (RF) to predict the label of wallpaper images, and the features used for predicting are Gabor feature and deep feature. We used the Pearson product-moment correlation coefficient to analyze the correlation between the predicted labels and the real labels, as shown in Fig.~\ref{fig:corr}. From left to right, from top to bottom, they are the real labels, the labels predicted by Gabor features with label distribution learning, the labels predicted by deep feature with label distribution learning, the labels predicted by Gabor features with regression Random Forest, and the deep feature with the label predicted by the regression Random Forest. The more similar the distribution of the predicted label and the real label, the higher the correlation between them. It can be seen from the experimental results that Gabor features and label distribution learning can achieve the best label prediction results.

\subsection{Wallpaper texture generation}
\begin{figure*}
\centering
  \includegraphics[width=0.97\textwidth]{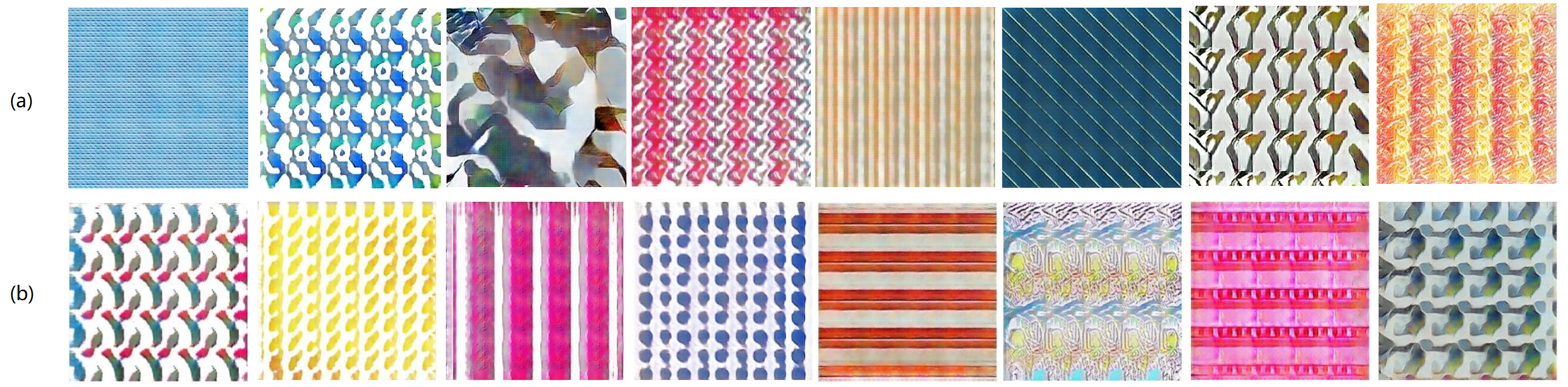}
\caption{Texture samples generated using proposed perception driven wallpaper generation model based on multi-label semantics, including repetitive, wavy, rhombus, marble, etc. (a) shows the wallpaper images generated with the multi-label semantics obtained from the semantic description collection survey, and (b) shows the wallpaper images generated with the multi-label semantics obtained from a multi-label learning method.}
\label{fig:generesults}       
\end{figure*}

\begin{figure*}
\centering
  \includegraphics[width=0.96\textwidth]{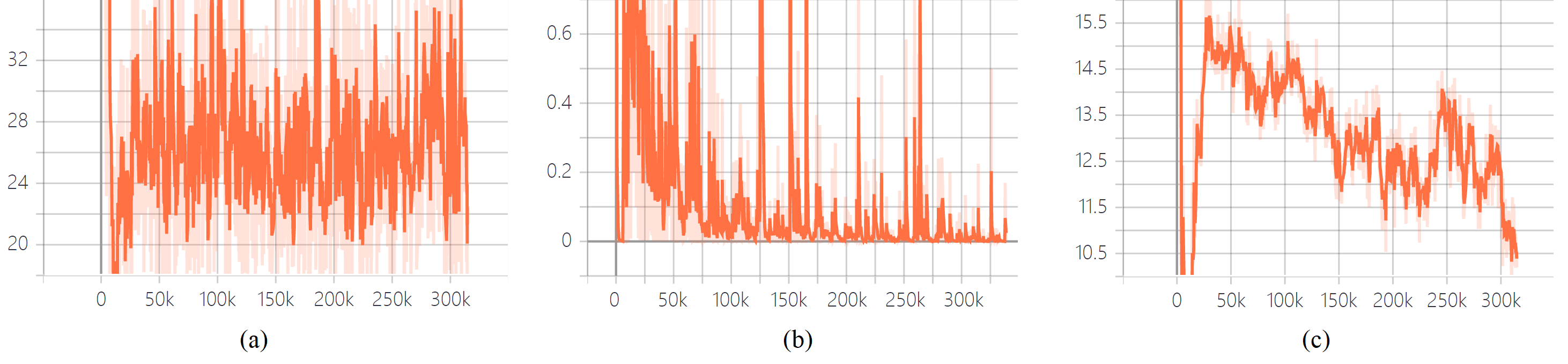}
\caption{The loss evolution for each part of the joint model as the number of iterations increases during training, (a) is the loss of the generator network; (b) is the loss of the discriminator network; (c) is the loss of the perceptual model.}
\label{fig:training}       
\end{figure*}

The experimental results of perception driven wallpaper texture generation are shown in Fig.~\ref{fig:generesults}. The output wallpaper textures are generated according to different semantic attributes, including repetitive, wooden, zig-zag, wavy, symmetrical, rhombus, spotted, marble, etc. The loss evolution for each part of the joint model during the training process, along with the number of iterations, is displayed in Fig.~\ref{fig:training}. From the figure, we can see that the loss of the generator network changes smoothly during the training process; the loss of the discriminator network becomes stable, which proves that our model has a good ability to generate wallpaper and discriminate the semantic descriptions of the generated wallpaper.

The proposed perception driven wallpaper texture generation network based on multi-label semantics can generate new wallpaper texture images according to semantic descriptions. The generated wallpaper images look clear and vary in styles. Their image size is $299\times299$, with a high resolution and rich texture details. Given a set of semantic labels, the trained model can generate 60-70 new wallpaper texture images per minute, based on multi-label semantics. This is simply achieved on RAM: 32G, CPU: 3.5GHz*8, GPU: 1080Ti 11G *2 workstation without further optimization.

\begin{figure*}
\centering
  \includegraphics[width=0.9\textwidth]{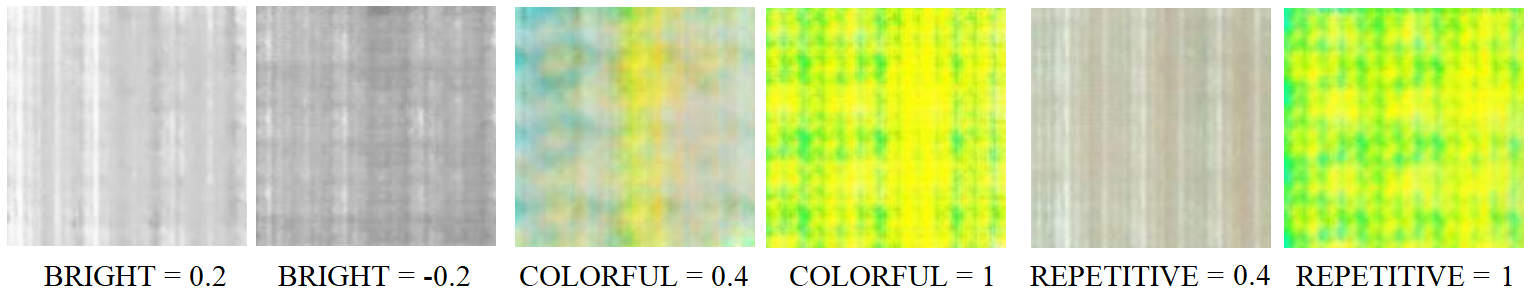}
\caption{Textures generated with certain semantic attribute values.The semantic attribute BRIGHT values of the first and second textures are 0.2 and -0.2, respectively; the semantic attribute COLORFUL values of the third and fourth textures are 0.4 and 1, respectively; the semantic attribute REPETITIVE values of the fifth and sixth textures are 0.4 and 1, respectively.}
\label{fig:change1}       
\end{figure*}

Given the semantic descriptions of wallpaper textures, the trained network is tested to generate wallpaper images that are consistent with the semantic descriptions. We fixed certain semantic attribute values, and then wallpaper textures change along with the other semantic attribute values. Fig.~\ref{fig:change1} shows the details of the textures generated according to the given semantic attribute values. Since most of the semantic attribute values are fixed when generating textures, the color of the generated textures are monotonous. The three representative semantic attributes are set to different values when generating textures. The semantic attribute BRIGHT values of the first and second textures are 0.2 and -0.2 respectively, the generated texture images have obvious differences in the bright attributes, the second texture is darker than the first texture; the semantic attribute COLORFUL values given to the third and fourth textures are 0.4 and 1, respectively, the larger the attribute value, the richer the texture color; the semantic attribute REPETITIVE values of the fifth and sixth textures are 0.4 and 1, respectively, the larger the attribute value, the more obvious the repeatability of the texture.

\subsection{Style transfer and aesthetic evaluation}
\begin{figure*}
\centering
  \includegraphics[width=0.98\textwidth]{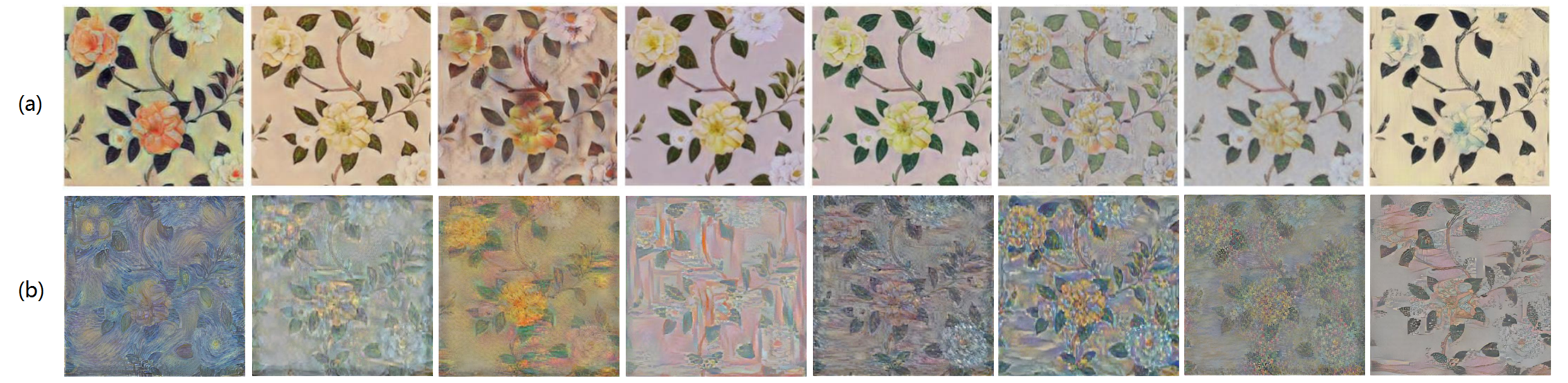}
\caption{The result of artistic style transfer of wallpaper texture images, (a) uses the CycleGAN for style transfer, and (b) uses the method proposed by Gatys et al. \cite{Gatys2015Neural}. From left to right, the generated wallpaper images are in the following order: Vincent Van Gogh, Paul Cezanne, Paul Gauguin, Henri Matisse, Pyotr Konchalovsky, Maurice Prendergast, Claude Monet and ukiyoe artists.}
\label{fig:style}       
\end{figure*}
The experimental results of style transfer are shown in Fig.~\ref{fig:style}. The generated wallpaper images are identical in semantics to the original image, but with a different hue and visual perception, typical of the new style. We added an aesthetic evaluation mechanism based on style transfer to evaluate the texture image and determine whether it conforms to most people aesthetics assessment.

The traditional image quality assessment mainly concerns the loss induced by image processing and displaying steps, so that the distortion degree of images can be measured. The current image aesthetic quality assessment evaluates the image using aesthetics factors such as illumination, composition of a picture, depth of field and color, in an attempt to simulate the human aesthetic ability. In 2017, the Google team proposed NIMA \cite{Talebi2017NIMA} based on deep convolutional neural network. This model does not only distinguish whether an image is appealing or not, or make a regression on the average of the scores, but it rather makes a score distribution of the images to be evaluated, whose value ranges in [1, 10]. The NIMA model assigns scores according to the score probability of these images. The trained model predicts a score very close to the score distribution from humans.
\begin{figure*}
\centering
  \includegraphics[width=0.98\textwidth]{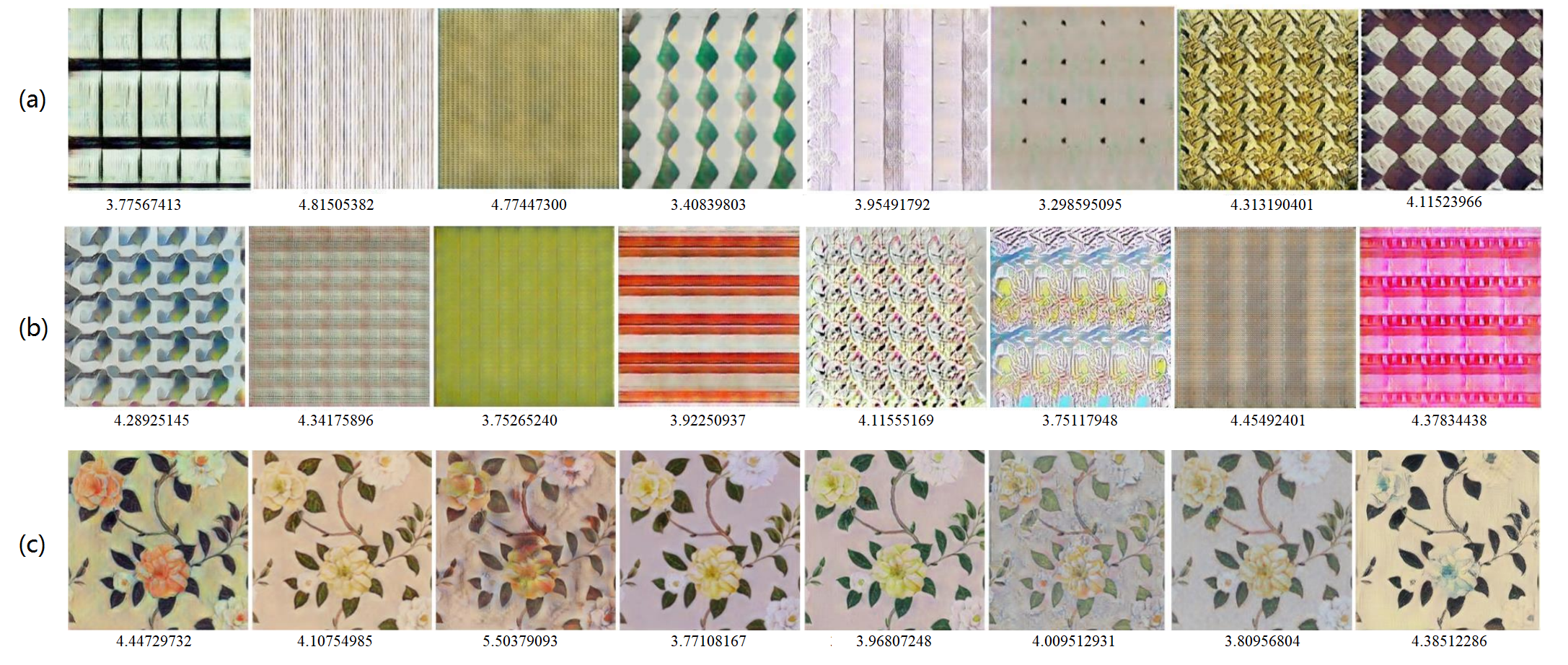}
\caption{Scoring results for aesthetic evaluation of generated wallpaper images using the NIMA algorithm. The wallpapers in (a) use the real semantic label distribution as the perceptual feature; the perceptual attributes of the wallpapers in (b) are the semantic label distributions predicted by the label distribution learning; and the wallpapers in (c) are style transfer results.}
\label{fig:score}       
\end{figure*}
We input both generated wallpaper images and those obtained from style transfer to the NIMA model for aesthetic evaluation. Fig.~\ref{fig:score} illustrates the experimental results, whose score below each wallpaper image is the average of the predicted aesthetic score distribution. The wallpapers in the first line are generated by using the real semantic label distribution as the perceptual feature to control the wallpaper texture generation process; the perceptual attributes of the wallpapers in the second line are the semantic label distributions predicted by the label distribution learning. In Fig.~\ref{fig:score}, when the wallpaper images are generated using the real label and the predicted label, their aesthetic scores are not much different. Therefore, in future wallpaper image processing studies, the predicted label distribution values can be used to expand our wallpaper texture dataset.

The wallpaper images produced by style transfer are shown in the third line of Fig.~\ref{fig:score}. The average value of the aesthetic score distribution may not be too consistent with human visual perception. We suspect this is because our style transfer model is trained on art works with artistic styles such as blurring, dim and sharp contrasts, hence cannot use the same aesthetic scoring rules as ordinary images. The average value of the score distributions for all wallpaper images is mainly concentrated in [3, 6]. After investigation, we suppose this is because NIMA is trained on the AVA dataset \cite{Murray2012AVA}, and the images in this dataset are high-definition photos. The aesthetic evaluation of these photos is mainly evaluated from the perspectives of color, composition of a photo, depth of field, brightness and contrast. In contrast, the wallpaper image is relatively simple and lacks fine-grained details contained in many of these photos. Overall, when using the NIMA model to evaluate generated wallpaper images based on multi-label semantics, most of the scores obtained are objective and consistent with human aesthetics. This method significantly saves manpower and time for human visual evaluation of wallpaper texture images, and has certain reference value and promising application prospect.

\section{Ablation Experiments}
\label{sec:Ablation}
\begin{figure*}
\centering
  \includegraphics[width=0.95\textwidth]{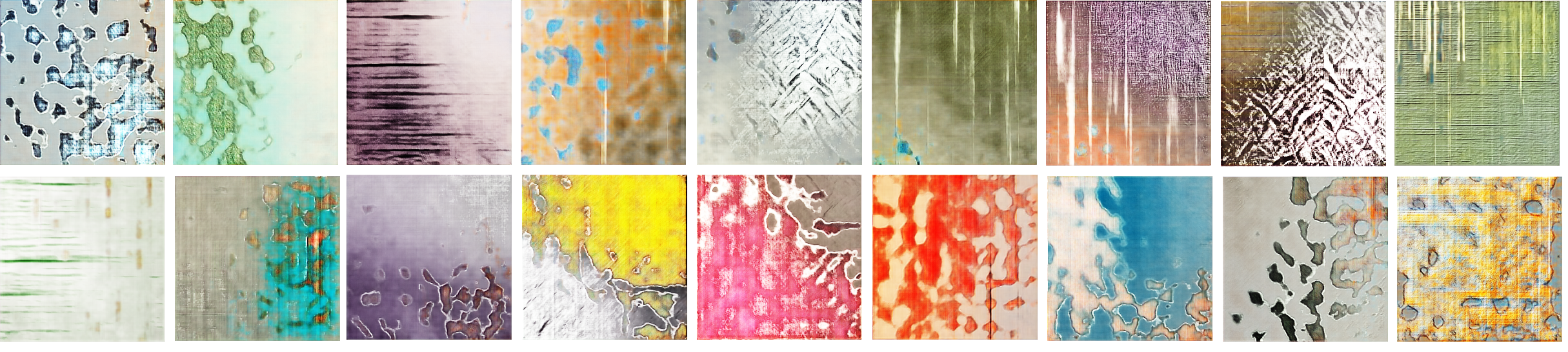}
\caption{The samples generated from DCGAN without semantic labels. The images are blur with low resolution and have a lot of noise.}
\label{fig:dcgan}       
\end{figure*}

It is known that Deep Convolutional Generative Adversarial Networks(DCGAN)\cite{Radford2015Unsupervised} is one of the representative model for image generation using Convolutional Neural Networks in deep learning. DCGAN is trained by setting a game between two models: generative model G and discriminative model D. G was trained to generate the samples which can deceive D, and the samples are intended to come from the same probability distribution as the training data (i.e. $p_{data}$), without having access to such data. D was trained to distinguish the samples from G rather than $p_{data}$. D and G play the two-player min-max game with the following objective function:
\begin{eqnarray}
\min_G \max_D V(D,G) &=& E_{x\sim p_{data}(x)}[\log D(x)]+\nonumber\\
                     & &E_{z\sim p_z(z)}[\log(1-D(G(z)))]
\end{eqnarray}
where $z$ is a noise vector from $p_z$, and $x$ is a real image from the data distribution $p_{data}$. DCGAN make some changes on the architecture of CNN. Therefore, it was more stable than GAN, enhanced the quality of samples and accelerated the speed of convergence. We use our wallpaper dataset without semantic labels to train the network, and a total of 299,600 iterations are performed. The generated wallpapers are show in Fig.~\ref{fig:dcgan}. Without the semantic labels, the results are not good enough. The generated images are blur with low resolution and have no evident change. These images only have obvious changes in color, and almost no useful wallpaper elements are learned in texture details. Besides, there are a lot of noise, and the quality of the generated images are not good enough. It can be seen that, without the driving of perceptual description, it is difficult to produce our expected images by directly using noise to generate wallpaper images. This proves that our perceptual model trained with multi-label semantic information can effectively drive the model to generate wallpaper images that meet human perception.

\begin{figure*}
\centering
  \includegraphics[width=0.95\textwidth]{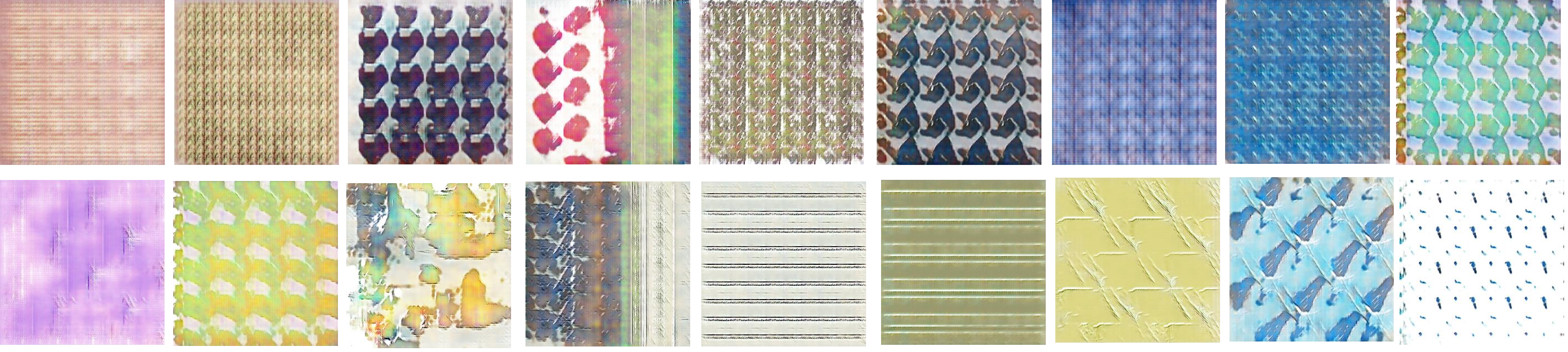}
\caption{The samples generated with VggNet based perceptual model. The images has obvious texture details, but not beautiful enough compared with the perceptual model based on Inception-V3.}
\label{fig:vgg}       
\end{figure*}

In order to further prove the effectiveness of the perceptual model, we replaced the perceptual model and conducted a controlled experiment. VggNet\cite{Kaiming2015Deep} is an excellent deep convolutional neural network structure, which has achieved good results in many classification tasks. Here we choose to use VggNet to replace Inception-V3 for pre-training the perceptual model. VggNet is originally used for image recognition, which is a classification task. In order to employ VggNet for perceptual feature regression, we also need to appropriately modify the network structure. The activation function of the output in the original network is modified into a hyperbolic tangent function, and the definition of the loss function is changed into quadratic loss. We pre-train the modified perceptual model, and then fix it as a part of the joint model. During the training process, a total of 210,000 iterations are performed, and the experimental results are shown in Fig.~\ref{fig:vgg}.

\begin{figure}
\centering
  \includegraphics[width=0.48\textwidth]{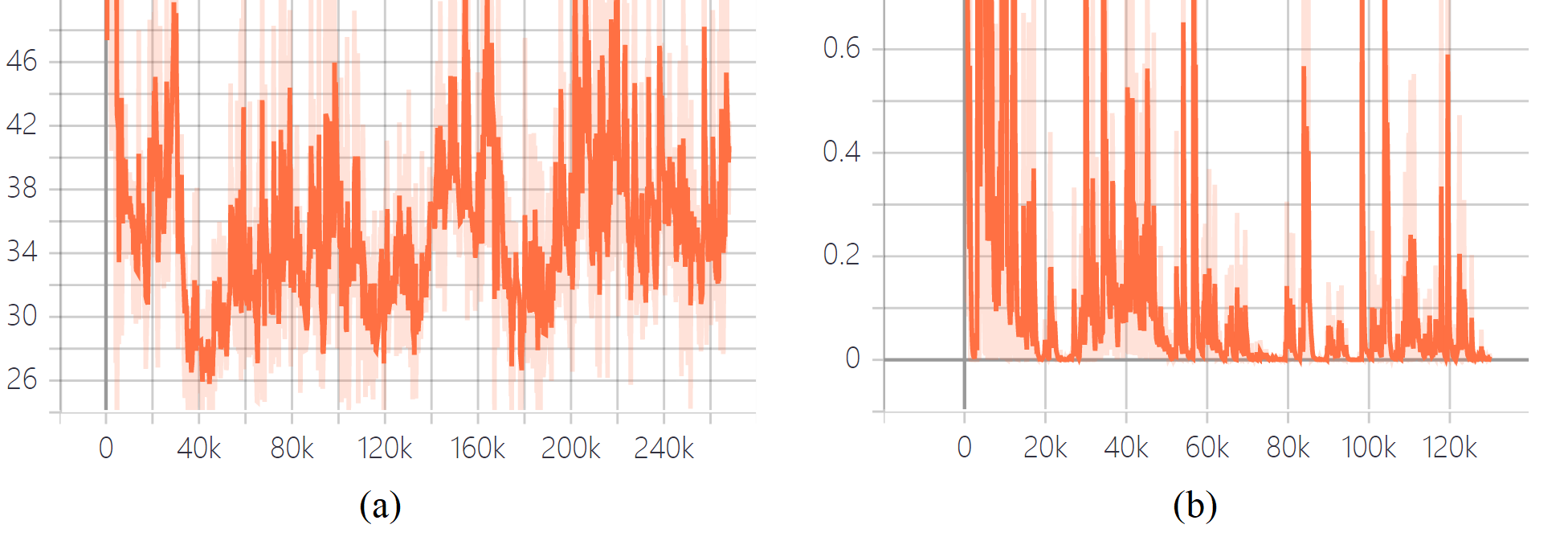}
\caption{The loss evolution during training the joint model with VggNet based perceptual model, (a) is the loss of the generator network;
(b) is the loss of the discriminator network.}
\label{fig:vgglog}       
\end{figure}

The generated image has obvious texture details, and the distribution of elements in the textures has regularity. However, compared with the perceptual model based on Inception-V3, the generated images are not vivid enough, and the appearance is not beautiful enough. In addition, VggNet has more parameters than Inception-V3, so it needs more computing resources. The loss evolution for each part of this new joint model during the training process is displayed in Fig.~\ref{fig:vgglog}. With the increase of the number of iterations, the loss of the discriminator network becomes gradually stabilized, but comparing with Fig.~\ref{fig:training}, the loss of the generator network still oscillates at a high level, which shows that when VggNet is used as the perceptual model, the ability of the joint model is not good enough to produce wallpaper images with a given semantic descriptions.

\begin{figure}
\centering
  \includegraphics[width=0.48\textwidth]{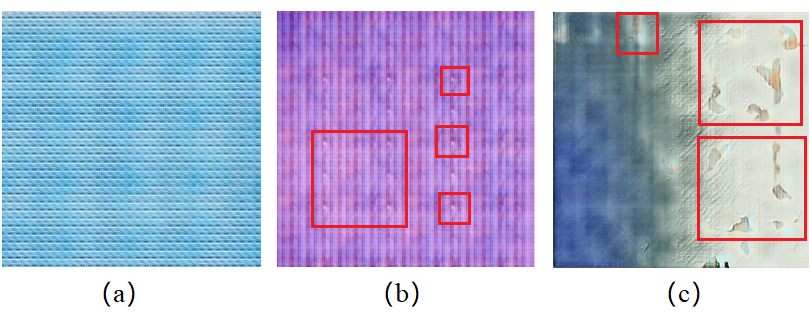}
\caption{Comparison of wallpaper images obtained according to different generation methods. (a) uses the Inception-V3 pre-trained perceptual model, (b) uses the VggNet pre-trained perceptual model, and (c) is directly generated using DCGAN.}
\label{fig:comp}       
\end{figure}

In order to compare the above three wallpaper texture image generation methods, we selected a representative set of generated results for comparison, as shown in the Fig .~\ref{fig:comp}. Fig.~\ref{fig:comp} (a) and (b) are the results produced from perception-driven texture generation method proposed in this paper. Among them, the perceptual model of (a) uses Inception-V3, and the perceptual model of (b) uses VggNet. Fig.~\ref{fig:comp} (c) is the generating result without semantic labels using the DCGAN method. It can be found that the wallpaper images generated by the perception-driven model have richer texture details, and the elements in the images are clearer and the color transition is uniform. At the same time, Fig.~\ref{fig:comp} (a) is relatively symmetrical, and there is no particularly abrupt noise, the noise in (b) is not very obvious, but in Fig.~\ref{fig:comp} (c), it can be seen that the noise and excessive color are very obvious. The red box marked the location of the obvious noise in the generated image.

\section{Conclusion}
\label{sec:conclusion}
The study of texture semantic is crucial in many research fields of computer vision. This paper analyzes the wallpaper texture image synthesis and generation and constructs a wallpaper texture dataset with semantic attributes. Based on multi-label semantics, using the generative adversarial network and perceptual feature regression, a perception driven wallpaper texture generation model is proposed, which can generate clear wallpaper texture images according to given semantic descriptions.

Based on this, we perform style transfer on the generated wallpaper images. Through the learning of artists' work style, a variety of wallpapers images with different styles are generated in real time. Aesthetic evaluation of wallpaper texture images by user survey consumes manpower and time. The generated wallpaper texture images need to be evaluated to make sure they are conform to human aesthetic assessment. We propose to use an aesthetic quality assessment method to predict the aesthetic score distribution of the generated wallpaper images, which is a very practical research.

There are still many issues to be solved in this paper. To bring more details into the generated wallpaper textures, a later work will be the optimization of the perception driven wallpaper texture generation model. From another perspective, aesthetic evaluation of textures and a more effective wallpaper generation can be designed to yield better results for the semantics out of our database. In addition, the proposed texture generation model can also be applied to generating animation texture elements, generating background textures for games, and generating materials, patterns and textures for other interior design, etc. Considering the resource-constrained architecture, we will use knowledge distillation to compress the model. Knowledge distillation essentially fixes the traditional generator as a teacher network, and designs a small network as a student network. These will be further studied in our future work.

\ifCLASSOPTIONcaptionsoff
  \newpage
\fi

\end{document}